# Non-monotonic Negation in Probabilistic Deductive Databases

**Raymond T. Ng** and **V.S. Subrahmanian**
Department of Computer Science
A. V. Williams Building
University of Maryland
College Park, Maryland 20742, U.S.A.

## Abstract

In this paper we study the uses and the semantics of non-monotonic negation in probabilistic deductive databases. Based on the stable semantics for classical logic programming, we introduce the notion of stable formula functions. We show that stable formula functions are minimal fixpoints of operators associated with probabilistic deductive databases with negation. Furthermore, since a probabilistic deductive database may not necessarily have a stable formula function, we provide a stable class semantics for such databases. Finally, we demonstrate that the proposed semantics can handle default reasoning naturally in the context of probabilistic deduction.

## 1 Introduction

Many frameworks on multivalued logic programming have been proposed to handle uncertain information, such as the ones described in [3, 9, 13, 14, 24]. However, all these approaches are non-probabilistic in nature, as the way they interpret conjunctions and disjunctions is too restrictive for probabilistic data. Since probability theory is well understood, we believe that a probabilistic approach to quantitative deduction in logic programming is important. In [16, 17] we have proposed a framework for probabilistic deductive databases, i.e. logic programs without function symbols. We show that this framework is expressive, as among others, it supports conditional probabilities, classical negation, propagation of probabilities, and Bayesian updates (cf. Example 1).

However, one fundamental issue that remains unaddressed in the framework proposed in [16, 17] is the representation and manipulation of non-monotonic modes of negation. In particular, the framework is incapable of default reasoning and drawing negative conclusions based on the *absence* of positive information. Thus, our focus in this paper is to study the uses and the semantics of non-monotonic negation in probabilistic logic programming.

The semantical approach we adopt is based on the stable semantics of (classical) logic programming with negation [12]. In a nutshell, the stable semantics for classical logic programming makes a "guess" as to the set of formulas provable from the program. Based on this guess, it transforms the program into a new program containing no occurrences of negation, and verifies if this guess satisfies some reasonable criterion. The "guess" is said to be *stable* if it satisfies the criterion.

In this paper we propose a stable semantics that is natural for probabilistic logic programs with non-monotonic negation. We introduce the notion of *stable formula functions*, and examine various connections between this notion of stability and fixpoints of operators associated with this kind of probabilistic logic programs. In effect, stable formula functions provides a fixpoint semantics for probabilistic logic programs with negation. However, similar to the situation in classical logic programming, not all probabilistic logic programs have stable formula functions. Based on the semantics proposed in [2] for classical logic programming, we thus provide a more general notion of *stable classes* of formula functions that applies to all probabilistic logic programs.

Section 2 presents the syntax and uses of probabilistic logic programs (without function symbols) with negation. Section 3 reviews the fixpoint theory for positive probabilistic logic programs presented in [16, 17]. Section 4 presents the notion of stable formula functions. It relates stable formula functions to fixpoints of operators associated with programs with negation. As programs may not have stable formula functions, Section 5 extends the notion of stability to provide semantics for such programs. Section 6 discusses how negation supports default reasoning, and compares our framework with related work. The last section concludes this paper with a discussion on future work.

## 2 Syntax and Uses of General Probabilistic Logic Programs

### 2.1 Gp-Clauses and Programs

Let $L$ be a language generated by finitely many constant and predicate symbols. While $L$ does *not* contain any ordinary function symbols, it may contain symbols for a fixed family of interpreted and computable[1] functions, known as *annotation functions* defined as follows.

**Definition 1** An *annotation function* $f$ of arity $n$ is a total function of type $([0,1])^n \rightarrow [0,1]$. □

We also assume that $L$ contains infinitely many variable symbols which are partitioned into two infinite subsets. The first subset consists of normal variable symbols in first order logic; they can only appear in atoms. We refer to these variables as *object variables*. The other set consists of *annotation variable* symbols. Annotation variables can only range between 0 and 1. Annotation variable symbols can only appear in *annotation terms*, a concept defined as follows.

**Definition 2** 1) $\rho$ is called an *annotation item* if it is one of the following:
i) a constant in [0,1], or
ii) an annotation variable in $L$, or
iii) of the form $f(\delta_1, \ldots, \delta_n)$, where $f$ is an annotation function of arity $n$ and $\delta_1, \ldots, \delta_n$ are annotation items.
2) For real numbers $c, d$ such that $0 \leq c, d \leq 1$, let the *closed interval* $[c,d]$ be the set $\{x \mid c \leq x \leq d\}$.
3) $[\rho_1, \rho_2]$ is called an *annotation (term)* if $\rho_i$ $(i = 1, 2)$ is an annotation item.
If an annotation does not contain any annotation variables, the annotation is called a *c-annotation*. □

Let $B_L$ denote the Herbrand base of $L$. Since $L$ does not contain any function symbols[2], $B_L$ is finite.

**Definition 3** 1) A *basic formula*, not necessarily ground, is either a conjunction or a disjunction of atoms. Note that both disjunction and conjunction cannot occur simultaneously in one basic formula.
2) Let $bf(B_L)$ denote the set of all ground basic formulas obtained by using distinct atoms in $B_L$, i.e. $bf(B_L) = \{A_1 \wedge \ldots \wedge A_n | n \geq 1$ is an integer and $A_1, \ldots, A_n \in B_L$ and $\forall 1 \leq i,j \leq n, i \neq j \Rightarrow A_i \neq A_j\}$ $\bigcup$ $\{A_1 \vee \ldots \vee A_n | n \geq 1$ is an integer and $A_1, \ldots, A_n \in B_L$ and $\forall 1 \leq i,j \leq n, i \neq j \Rightarrow A_i \neq A_j\}$, where all $A_i$'s are ground atoms. □

[1] A function $f$ is computable in the sense that there is a fixed procedure $P_f$ such that if $f$ is $n$-ary, and $\mu_1, \ldots, \mu_n$ are given as inputs to $P_f$, then $f(\mu_1, \ldots, \mu_n)$ is computed by $P_f$ in a finite amount of time.

[2] Whenever we say function symbols, we mean exclusively the function symbols in normal first order logic, not including the annotation function symbols defined previously.

**Definition 4** 1) Let $F_0, \ldots, F_n, G_1, \ldots, G_m$ be basic formulas. Also let $\mu_0, \ldots, \mu_{n+m}$ be annotations such that every annotation variable occurring in $\mu_0$, if any, also appears in one of $\mu_1, \ldots, \mu_{n+m}$. Then the clause

$$F_0 : \mu_0 \leftarrow F_1 : \mu_1 \wedge \ldots \wedge F_n : \mu_n \wedge \neg(G_1 : \mu_{n+1}) \wedge \ldots \wedge \neg(G_m : \mu_{n+m})$$

is called a *general probabilistic clause* (*gp-clause* for short).
2) A *pf-clause* is a gp-clause without negated annotated basic formulas, i.e. $m = 0$ [17]. □

**Definition 5** 1) A *general probabilistic (gp-)program* is a finite set of gp-clauses.
2) A *pf-program* is a finite set of pf-clauses [17]. □

If the annotation $\mu$ is a c-annotation $[c_1, c_2]$, the annotated basic formula $F : \mu$ intuitively means: "The probability of $F$ must lie in the interval $[c_1, c_2]$." Similarly, the negation of the annotated formula, $\neg(G : \mu)$, is to be read as: "It is not provable that the probability of $G$ must lie in the interval $\mu$." Hence, the negation $\neg$ considered here is non-monotonic. Finally, note that to specify that the probability of $F$ is a point $c$, simply use $F : [c, c]$. The reason why we prefer to support probability ranges to probability points is that given the probabilities of two formulas $F_1, F_2$, it is generally not possible to *precisely* state the probabilities of $(F_1 \wedge F_2)$ and $(F_1 \vee F_2)$ [16]. It is however possible to precisely state the tightest range within which these probabilities must lie. In [8] Fagin and Halpern also propose using an interval to represent the degree of belief for a nonmeasurable event.

### 2.2 Uses of Gp-programs

In the following we show a few examples to demonstrate the expressive power of gp-clauses. More examples on default reasoning are included in Section 6.

**Example 1** In [17, 18], we show how to use gp-clauses to support propagation of probabilities, classical negation and van Emden's quantitative rule processing[24]. Due to space limitations, here we only show how to support conditional probabilities and Bayesian updates in our framework.
i) (**Support for Conditional Probabilities**:) Suppose the conditional probability of $A$ given $B$ is known to be $p$. This is equivalent to saying: $Prob(A \wedge B) = p * Prob(B)$. Thus, we can use the pf-clause:

$$(A \wedge B) : [p * V_1, p * V_1] \leftarrow B : [V_1, V_1].$$

Similarly, if to calculate the conditional probability of $A$ given $B$, denoted by $(A|B)$, we can use:

$$(A|B) : [V_2/V_1, V_2/V_1] \leftarrow (A \wedge B) : [V_2, V_2] \wedge B : [V_1, V_1],$$

assuming that the probability of $B$ is not 0. In [17] we show that these clauses maintain the intended conditional probability relationships.

iii) (**Bayesian Updates**:) Bayes rule states that: $Prob(B|A) = Prob(A|B) * Prob(B) / Prob(A)$ assuming $Prob(A) \neq 0$. Hence, the (updated) conditional probability of $B$ given $A$, can be calculated by the pf-clause:

$$(B|A) : [V_1 * V_2/V_3, V_1 * V_2/V_3] \leftarrow (A|B) : [V_1, V_1] \wedge B : [V_2, V_2] \wedge A : [V_3, V_3].$$

□

**Example 2** Suppose we believe that if it is not provable that the probability of a coin $C$ showing heads is within the range [0.49, 0.51], then there is over 95% chance that the coin is unfair. The gp-clause:

$$unfair(C) : [0.95, 1] \leftarrow \neg(head(C) : [0.49, 0.51])$$

represents our belief. □

**Example 3** Suppose we know that there is over 95% chance that a dog can bark, unless the dog is abnormal[3]. We also know that Benjy and Fido are dogs. However, Benjy is unable to bark (his vocal cords were injured at some point). This can be represented as:

$$\begin{aligned} bark(X) : [0.95, 1] &\leftarrow dog(X) : [1, 1] \wedge \neg(abn(X) : [1, 1]) \\ dog(fido) : [1, 1] &\leftarrow \\ dog(benjy) : [1, 1] &\leftarrow \\ bark(benjy) : [0, 0] &\leftarrow \\ abn(X) : [1, 1] &\leftarrow bark(X) : [0, 0] \end{aligned}$$

The last clause says that a dog is certainly abnormal if it *definitely* cannot bark. As we shall see later on (cf. Example 6), we can deduce from these clauses the fact that Fido can bark, but Benjy cannot. □

In [18] we also show how our framework can support mutual exclusion. See [18] for more details. In [17] we propose a fixpoint theory for pf-programs – negation-free gp-programs. Our objective here is to investigate how to extend this theory to handle negation. To do so, we adopt the stable semantical approach[12] for classical logic programming. But before we describe the stable semantics for gp-programs, we review the fixpoint theory we developed for pf-programs.

## 3 Background: Fixpoint Theory for Pf-programs

In this section we summarize the essential notions and results of the fixpoint theory we developed for probabilistic logic programs without negation as described in [16, 17]. Readers familiar with [16, 17] may skip this section.

[3] Note that this statement is *not* the same as saying: "Over 95% of all dogs bark."

**Definition 6** 1) Let a *world* $W$ be an Herbrand interpretation, i.e. a subset of $B_L$. For ease of presentation, assume there is an arbitrary, but fixed enumeration of all possible worlds/subsets of $B_L$. Such enumerations are possible as $L$ contains no function symbols.
2) A *world probability density function* $WP : 2^{B_L} \longrightarrow [0, 1]$ assigns to each world $W_j \in 2^{B_L}$ a probability $WP(W_j)$ such that for all $W_j \in 2^{B_L}$, $WP(W_j) \geq 0$ and $\sum_{W_j \in 2^{B_L}} WP(W_j) = 1$.
3) To simplify our notation, hereafter we use $p_j$ to denote $WP(W_j)$ for $W_j \in 2^{B_L}$. □

In the context of probabilistic deduction, we assume that the "real" world is definite, i.e. there are some propositions that are true, and some that are false. However, we are uncertain which of the various "possible worlds" is the right one. Thus, we use a world probability density function to define probability densities on the set of all possible worlds. In other words, a world probability density function assigns a probability (i.e. a non-negative number) to each world such that the sum of all probabilities adds up to 1. Our notions of worlds and world probability density functions are similar in essence to the "possible worlds" approach suggested by Nilsson [19]. While Nilsson's enumeration of the possible worlds is based on the given *set of sentences*, ours is based on the Herbrand interpretations of $L$.

In the study of the semantics of pf-programs, our aim is to use the probability ranges described in a pf-program to find the probabilistic truth values (i.e. point probabilities) of basic formulas. In particular, we use the probability ranges to find world probability density functions that obey those ranges. While the process will be formalized shortly, the following notion of a *formula function* is crucial for the process.

**Definition 7** 1) Let $\mathcal{C}[0, 1]$ denote the set of all closed sub-intervals of the unit interval [0,1], i.e. the set of all (contiguous) closed intervals $[c, d]$ that are subsets of [0, 1].
2) A *formula function* is a mapping $h : bf(B_L) \longrightarrow \mathcal{C}[0, 1]$. □

The empty interval, denoted by $\emptyset$, is a member of $\mathcal{C}[0, 1]$, because it may be represented as $[c_1, c_2]$ where $c_2 < c_1$. Intuitively, a formula function assigns a probability range to each ground basic formula. Then given a formula function, we can find world probability density functions that obey the ranges assigned by the formula function. This is achieved by setting up a set of linear constraints, as described in the following definition.

**Definition 8** 1) Let $h$ be a formula function. A set of linear constraints, denoted by $\mathcal{LC}(h)$, is defined as follows. For all $F_i \in bf(B_L)$, if $h(F_i) = [c_i, d_i]$, then the inequality $c_i \leq \left( \sum_{W_j \models F_i \text{ and } W_j \in 2^{B_L}} p_j \right) \leq d_i$ is in $\mathcal{LC}(h)$ (where $p_j$'s are used as specified in Definition 6). In addition, $\mathcal{LC}(h)$ contains the following 2 constraints: $\sum_{W_j \in 2^{B_L}} p_j = 1$ and $(\forall W_j \in 2^{B_L}), p_j \geq 0$.

2) Let $\mathcal{WP}(h)$ denote the solution set of $\mathcal{LC}(h)$. □

It is easy to see that each solution $WP \in \mathcal{WP}(h)$ (i.e. the solution set of $\mathcal{LC}(h)$) is a world probability density function. Also note that $2^{B_L}$ consists of all possible worlds, and any two distinct worlds are mutually incompatible as they must differ on at least one atom. Thus, given a world probability density function $WP$, we can compute the probabilistic truth value of any basic formula $F$ with respect to $WP$ by adding up the probabilities of all the possible worlds in which $F$ is true in the classical 2-valued sense. Hence, it is the set $\mathcal{LC}(h)$ of linear constraints that enables us to find probabilistic truth values that satisfy the ranges assigned by the formula function $h$. Now we are in a position to define a fixpoint operator $T_P$ for pf-programs $P$. Hereafter we use the notation $\mathcal{FF}$ to denote the set of all formula functions, and $\mathbf{min}_{\mathbf{Q}}(Exp)$ and $\mathbf{max}_{\mathbf{Q}}(Exp)$ to denote the minimization and maximization of the expression $Exp$ subject to the set of constraints $\mathbf{Q}$.

**Definition 9** Suppose $P$ is a pf-program and $h$ is a formula function.
1) Define an intermediate operator $S_P : \mathcal{FF} \longrightarrow \mathcal{FF}$ as follows:
For all $F \in bf(B_L)$, $S_P(h)(F) = \bigcap M_F$ where $M_F = \{\alpha \mid F : \alpha \leftarrow F_1 : \alpha_1 \wedge \ldots \wedge F_n : \alpha_n$ is a ground instance of a clause in $P$, and for all $1 \leq i \leq n, h(F_i) \subseteq \alpha_i\}$. In particular, if $M_F$ is empty, set $S_P(h)(F) = [0, 1]$.

2) Define $T_P : \mathcal{FF} \longrightarrow \mathcal{FF}$ as follows:
i) If $\mathcal{WP}(S_P(h))$ is non-empty (i.e. $\mathcal{LC}(S_P(h))$ has solutions), then for all $F \in bf(B_L)$, $T_P(h)(F) = [c_F, d_F]$ where

$$c_F = \mathbf{min}_{\mathcal{LC}(\mathbf{S_P(h)})} \left( \sum_{W_j \models F \text{ and } W_j \in 2^{B_L}} p_j \right) \text{ and}$$

$$d_F = \mathbf{max}_{\mathcal{LC}(\mathbf{S_P(h)})} \left( \sum_{W_j \models F \text{ and } W_j \in 2^{B_L}} p_j \right).$$

ii) Otherwise, if $\mathcal{WP}(S_P(h))$ is empty, then for all $F \in bf(B_L)$, $T_P(h)(F) = \emptyset$. □

Informally, $S_P(h)$ is a one-step immediate consequence operator that determines the probability ranges of basic formulas by one-step deductions of the pf-clauses in $P$. But since basic formulas can appear as the heads of pf-clauses, as an example the following situation may arise: $S_P(h)(A \vee B) = [0, 0]$, but $S_P(h)(A) = S_P(h)(B) = [1, 1]$. By regarding $[1, 1]$ as true and $[0, 0]$ as false, these range assignments are not consistent. In general, "local" assignments of probability ranges to formulas may not be "globally" consistent. Hence, the linear program $\mathcal{LC}(S_P(h))$ is set up to ensure that all assignments are consistent. Then $T_P$ assigns to each formula a probability range that satisfies every constraint in the linear program.

Given two formula functions $h_1$ and $h_2$, we say that $h_1 \leq h_2$ iff $\forall F \in bf(B_L), h_1(F) \supseteq h_2(F)$. As shown in [16], the set $\mathcal{FF}$ of formula functions forms a complete lattice with respect to the ordering $\leq$ defined above. Moreover, the $\top$ element is the formula function $h$ such that $\forall F \in bf(B_L), h(F) = \emptyset$, and the $\perp$ element is the one such that $\forall F \in bf(B_L), h(F) = [0, 1]$. In [17] we show that $T_P$ is monontonic, and thus there exists a least fixpoint $lfp(T_P)$ of $T_P$.

## 4 Stability of Formula Functions

In the presence of negation, the fixpoint operator associated with a pf-program (cf. Definition 9) must be extended to handle negation. We use $T'_P$ to denote the fixpoint operator associated with a gp-program $P$.

**Definition 10** Suppose $P$ is a gp-program and $h$ is a formula function.
1) Define an intermediate operator $S'_P : \mathcal{FF} \longrightarrow \mathcal{FF}$ as follows:
For all $F \in bf(B_L)$, $S'_P(h)(F) = \bigcap M'_F$ where $M'_F = \{\alpha \mid F : \alpha \leftarrow F_1 : \alpha_1 \wedge \ldots \wedge F_n : \alpha_n \wedge \neg(G_1 : \beta_1) \wedge \ldots \wedge \neg(G_m : \beta_m)$ is a ground instance of a clause in $P$, for all $1 \leq i \leq n, h(F_i) \subseteq \alpha_i$, and for all $1 \leq j \leq m, h(G_j) \not\subseteq \beta_j\}$. In particular, if $M'_F$ is empty, set $S'_P(h)(F) = [0, 1]$.
2) $T'_P$ is obtained from $S'_P$ in exactly the same way as $T_P$ is obtained from $S_P$. □

The example below shows that $T'_P$ is not monotonic.

**Example 4** Consider the gp-program $P$ described in Example 2:

$$p : [0.95, 1] \leftarrow \neg(q : [0.49, 0.51]).$$

Suppose $h_1$ is a formula function that assigns [0,1] to $q$, and $h_2$ is one that assigns [0.5,0.5] to $q$. Suppose that $h_1$ and $h_2$ assign [0,1] to all other basic formulas. Thus, it is the case that $h_1 \leq h_2$. But then, $T'_P(h_1)$ assigns [0.95,1] to $p$, while $T'_P(h_2)$ assigns [0,1] to $p$. Therefore, $T'_P(h_1)$ is not necessarily less than or equal to $T'_P(h_2)$. □

In the following we define the notion of *stable formula functions*, adapted from the stable model semantics proposed by Gelfond and Lifschitz [12]. We ultimately show that if there exists a stable formula function with respect to a gp-program $P$, the formula function is a minimal fixpoint of $T'_P$.

**Definition 11** Given a gp-program $P$ and a formula function $h$, the ***formula-function-transform (ff-transform*** for short) of $P$ based on $h$, denoted by $ff(P, h)$, is defined as follows:
1) Given a ground instance $C' \equiv F : \alpha \leftarrow F_1 : \alpha_1 \wedge \ldots \wedge F_n : \alpha_n \wedge \neg(G_1 : \beta_1) \wedge \ldots \wedge \neg(G_m : \beta_m)$ of a clause in $P$, if $h(G_j) \not\subseteq \beta_j$ for all $1 \leq j \leq m$, then the clause $C \equiv F : \alpha \leftarrow F_1 : \alpha_1 \wedge \ldots \wedge F_n : \alpha_n$ is included in $ff(P, h)$.
2) Nothing else is in $ff(P, h)$. □

**Definition 12** Let $P$ be a gp-program. A formula function $h$ is ***stable*** with respect to $P$ if $h$ is equal to the least fixpoint of $T_{ff(P,h)}$, i.e. $h = lfp(T_{ff(P,h)})$. □

**Example 5** Consider again the gp-program $P$:

$$p : [0.95, 1] \leftarrow \neg(q : [0.49, 0.51]),$$

where $p, q$ are ground. Then given the formula function $h_1$ such that $h_1(p) = [0.95, 1]$ and $h_1(q) = [0, 1]$, the ff-transform of $P$ based on $h_1$ is the single clause:

$$p : [0.95, 1] \leftarrow .$$

Then the least fixpoint $lfp(T_{ff(P,h_1)})$ assigns [0.95, 1] to $p$ and [0,1] to $q$. Hence, $h_1$ is stable. In fact, it is easy to show that $h_1$ is the only stable formula function[18]. □

**Example 6** Consider the gp-program $P$ for Benjy and Fido shown in Example 3. Let a formula function $h_1$ assigns [0,0] to $bark(benjy)$, [1,1] to $abn(benjy)$, [0.95,1] to $bark(fido)$, and [0,1] to $abn(fido)$. Then $ff(P, h_1)$ consists of the following clauses:

$$\begin{aligned} bark(fido) : [0.95, 1] &\leftarrow dog(fido) : [1, 1] \\ dog(fido) : [1, 1] &\leftarrow \\ dog(benjy) : [1, 1] &\leftarrow \\ bark(benjy) : [0, 0] &\leftarrow \\ abn(benjy) : [1, 1] &\leftarrow bark(benjy) : [0, 0] \\ abn(fido) : [1, 1] &\leftarrow bark(fido) : [0, 0]. \end{aligned}$$

It is easy to check that $h_1 = lfp(T_{ff(P,h_1)})$. Therefore, $h_1$ is stable. In fact, it is easy to verify that $h_1$ is the only stable formula function for this gp-program. □

The examples below show that there are gp-programs that have none or more than one stable formula function.

**Example 7** The gp-program that consists of the single clause below:

$$p : [0.95, 1] \leftarrow \neg(p : [0.95, 1])$$

does not have a stable formula function. See [18] for a proof. □

**Example 8** The gp-program that consists of the following clauses:

$$\begin{aligned} p : [0.95, 1] &\leftarrow \neg(q : [0.49, 0.51]) \\ q : [0.49, 0.51] &\leftarrow \neg(p : [0.95, 1]) \end{aligned}$$

has two stable formula functions: i) $h_1$ such that $h_1(p) = [0.95, 1]$ and $h_1(q) = [0, 1]$, and ii) $h_2$ such that $h_2(p) = [0, 1]$ and $h_2(q) = [0.49, 0.51]$. □

Intuitively, a stable formula function with respect to a gp-program makes "reasonable" guesses on the probability ranges assigned by the program to basic formulas. In particular, the following theorem shows that a stable formula function with respect to gp-program $P$ is a minimal fixpoint of $T'_P$ (cf. Definition 10).

**Theorem 1** Let $h$ be a stable formula function with respect to gp-program $P$. Then: $h$ is a minimal fixpoint of $T'_P$, i.e. there does not exist any formula function $h' < h$ such that $T'_P(h') = h'$. □

From Theorem 1, we can conclude that every stable formula function is a minimal fixpoint of $T'_P$. But the following example shows that the converse is not true.

**Example 9** Consider the gp-program $P$ that consists of the following clauses:

$$\begin{aligned} p : [0.95, 1] &\leftarrow \neg(p : [0.95, 1]) \\ p : [0.95, 1] &\leftarrow q : [1, 1] \\ q : [1, 1] &\leftarrow q : [1, 1]. \end{aligned}$$

It is easy to check that the formula function $h$ that assigns [0.95,1] to $p$ and [1,1] to $q$ is a minimal fixpoint of $T'_P$. However, $h$ is not stable [18]. □

Thus far, we have introduced the notion of stable formula functions which has the desirable property that it is a minimal fixpoint of $T'_P$. In effect, stable formula functions provide a fixpoint semantics for gp-programs. However, as shown in Example 7, a gp-program does not necessarily have a stable formula function. It is therefore the purpose of the next section to extend our theory of stability to cover those programs.

## 5 Stable Classes of Formula Functions

In [2] Baral and Subrahmanian propose a stable and extension class theory for logic programs and default logics. Here we adopt an analogous approach in proposing a stable class of formula functions defined as follows.

**Definition 13** Let $P$ be a gp-program, and $SF$ be a finite set of formula functions. Then: $SF$ is a ***stable class*** of formula functions with respect to $P$ iff $SF = \{lfp(T_{ff(P,h_i)}) \mid h_i \in SF\}$. □

Intuitively, a formula function $h_i$ in a stable class is the same as the least fixpoint of an operator associated with the ff-transform of $P$ based on some member $h_j$ in the stable class, i.e. $h_i = lfp(T_{ff(P,h_j)})$. In general, every member in the class is related in the same way with some other member in the class. See [2] for more details on stable class theory. In short, a stable class of formula functions with respect to a gp-program represents a set of "reasonable" guesses on the probability ranges assigned by the program to basic formulas.

**Example 10** Consider the gp-program in Example 7 again. A stable class of the program consists of the two formula functions: $h_1(p) = [0.95, 1]$ and $h_2(p) = [0, 1]$. It is easy to check that $h_1 = lfp(T_{ff(P,h_2)})$ and $h_2 = lfp(T_{ff(P,h_1)})$. □

**Lemma 1** A formula function $h$ is a stable formula function with respect to gp-program $P$ iff the singleton set $\{h\}$ is a stable class with respect to $P$. □

The aim of the remainder of this section is to prove that every gp-program has a non-empty stable class of formula functions (cf. Theorem 2).

**Definition 14** Let $P$ be a gp-program and $h$ be a formula function. Define the operator $\mathcal{SF}_P : \mathcal{FF} \longrightarrow \mathcal{FF}$ as: $\mathcal{SF}_P(h) = lfp(T_{ff(P,h)})$. □

**Lemma 2** The operator $\mathcal{SF}_P$ is anti-monotonic, i.e. $h_1 \leq h_2$ implies $\mathcal{SF}_P(h_2) \leq \mathcal{SF}_P(h_1)$. □

The following theorem is now an immediate consequence of the above lemma and a theorem by Yablo[25] and Fitting[10].

**Theorem 2** Every gp-program has a non-empty stable class of formula functions. □

The theorem above states that every gp-program has a non-empty stable class of formula functions. Suppose $C_1, C_2$ are two sets of formula functions. Recall that the ordering $\leq$ applies to formula functions. We extend this ordering now to *sets* of formula functions (and hence to stable classes) in two ways. Both orderings are well known in algebraic structures called *power domains* due to Hoare and Smyth [23].

**Definition 15** Let $S_1, S_2$ be two sets. We say that:
1) $S_1 \leq_{\mathbf{smyth}} S_2$ iff $(\forall s_1 \in S_1)(\exists s_2 \in S_2)\, s_1 \leq s_2$, and
2) $S_1 \leq_{\mathbf{hoare}} S_2$ iff $(\forall s_2 \in S_2)(\exists s_1 \in S_1)\, s_1 \leq s_2$. □

**Definition 16** 1) A non-empty stable class $C$ is said to be *Hoare-minimal* iff:
i) $C$ is inclusion-minimal, i.e. there is no non-empty stable class $C'$ such that $C' \subset C$ and
ii) for every inclusion-minimal non-empty finite stable class $C'$, $C' \leq_{\mathbf{hoare}} C$ implies $C' = C$.
2) $C$ is said to be *Smyth-minimal* iff condition (i) above holds and condition (ii) holds with $\leq_{\mathbf{hoare}}$ replaced by $\leq_{\mathbf{smyth}}$. □

We may choose either Hoare-minimal stable classes or Smyth-minimal stable classes as the intended meaning of our program. However, depending on the choice we make, we may get different semantics as shown below.

**Example 11** Consider the gp-program:

$$\begin{aligned} p : [1,1] &\leftarrow a : [1,1] \\ p : [1,1] &\leftarrow b : [1,1] \\ a : [1,1] &\leftarrow \neg(b : [1,1]) \\ b : [1,1] &\leftarrow \neg(a : [1,1]). \end{aligned}$$

This program has two stable formula functions: i) $h_1$ that assigns $[1,1]$ to both $p$ and $a$, and ii) $h_2$ that assigns $[1,1]$ to both $p$ and $b$. Furthermore, suppose $h_3$ is the function that assigns $[1,1]$ to all of $p, a, b$ and $h_4$ is the function that assigns $[0,1]$ to all of $p, a, b$. Then the set $\{h_3, h_4\}$ is a stable class of formula functions.

Note that here $\{h_1\}$ and $\{h_2\}$ are both Smyth-minimal stable classes of formula functions. Hence, the Smyth-minimal stable class semantics assigns $[1,1]$ to $p$. However, $\{h_3, h_4\}$ is the unique Hoare-minimal stable class of formula functions. This Hoare-minimal class only allows us to conclude that $p$ gets the value $[0,1]$. □

In short, we have introduced the notion of a stable class of formula functions for gp-programs. Lemma 1 shows that if $h$ is a stable formula function, then the singleton set $\{h\}$ is a stable class. Thus, the stable class semantics is defined for all gp-programs - whether or not they have stable formula functions.

## 6 Discussion

Like many researchers, we are interested in the use of numerical estimates in default reasoning. Unfortunately the framework we proposed in [17] is not powerful enough to handle default rules and exceptions. But now with the support of the non-monotonic negation $\neg$, we can specify that a default rule is only applicable in the absence of evidence to the contrary. Example 6 shows that the stable semantics proposed here handles the interaction between default rules and exceptions appropriately. Furthermore, the following examples demonstrate that the proposed semantics can also deal with interacting default rules.

**Example 12** In [21], Reiter and Criscuolo consider the following situation: i) that John is a high school dropout, ii) that high school dropouts are typically adults, and iii) that adults are typically employed. Due to transitivity of default rules ii) and iii), the conclusion that John is employed can be deduced. They argue that this conclusion is undesirable.

Now consider the following gp-program $P_1$:

$$\begin{aligned} adult(X) : [0.95, 1] &\leftarrow dropout(X) : [1,1] \\ employed(X) : [0.95, 1] &\leftarrow adult(X) : [1,1] \wedge \\ &\quad\ \neg(abn(X) : [1,1]) \\ abn(X) : [1,1] &\leftarrow dropout(X) : [1,1]. \end{aligned}$$

Suppose initially $P_1$ contains the fact: *adult(john)* : $[1, 1] \leftarrow$. Then it is easy to check that the only stable formula function with respect to $P_1$ assigns the range [0,95,1] to *employed(john)* correctly.

Suppose *dropout(john)* : $[1, 1] \leftarrow$ is added to $P_1$. Call this new program $P_2$. Consider the formula function $h$ that assigns [1,1] to *adult(john)*, *dropout(john)* and *abn(john)*, but [0,1] to *employed(john)*. It is easy to check that $h$ is a unique stable formula function with respect to $P_2$.

Finally, consider the situation where the only known fact about John is that he is a high school dropout, i.e. deleting the fact about John's adulthood from program $P_2$. Call this new program $P_3$. The unique stable formula function with respect to $P_3$ is the one that assigns: [1,1] to *dropout(john)* and *abn(john)*, [0.95,1] to *adult(john)*, and [0,1] to *employed(john)*. Hence, undesirable conclusions due to transitivity of default rules are avoided. □

The framework proposed by Dubois and Prade [6] also handles the situation discussed in the above example. However, their semantics is different from ours, as their framework is based on possibility logic and their model theory is based on fuzzy sets [26] which are well-known to be non-probabilistic. The following example on interacting default rules has been discussed extensively, but see [11, 20] for a probabilistic treatment on the subject.

**Example 13** Consider the situation: i) that tweety is a penguin, ii) that a penguin is a bird, iii) that typically penguins cannot fly, and iv) that birds can typically fly. The situation can be represented by the following gp-program:

$$\begin{aligned}
fly(X) : [0.95, 1] &\leftarrow bird(X) : [1, 1] \wedge \\
&\quad\; \neg(abnBird(X) : [1, 1]) \\
fly(X) : [0, 0.05] &\leftarrow penguin(X) : [1, 1] \wedge \\
&\quad\; \neg(abnPeng(X) : [1, 1]) \\
bird(X) : [1, 1] &\leftarrow penguin(X) : [1, 1] \\
abnBird(X) : [1, 1] &\leftarrow penguin(X) : [1, 1] \\
penguin(tweety) : [1, 1] &\leftarrow .
\end{aligned}$$

Consider the formula function $h$ that assigns: [1,1] to *penguin(tweety)*, *bird(tweety)* and *abnBird(tweety)*, [0,0.05] to *fly(tweety)*, and [0,1] to *abnPeng(tweety)*. Again it is easy to show that $h$ is the unique stable formula function with respect to the program. □

Thus far, we have shown several examples on how to handle default reasoning in our framework. But our framework is not as expressive as the probabilistic frameworks proposed by Bacchus[1] and Buntine[4]. For instance, given the above example, their frameworks can conclude that "birds typically are not penguins." Such a conclusion is not deducible in our framework, and in ongoing research we are studying how to extend our theory to handle such cases. However, as the framework of Bacchus extends full first-order logic, it is unclear to us how his framework can be used as a basis for logic programming and deductive databases. Similar comments apply to Buntine's proposal.

There have also been many proposals on multivalued logic programming. These include the works by Blair and Subrahmanian [3], Fitting [9], Kifer et al [13, 14], and van Emden [24]. However, they do not support non-monotonic modes of negation. On the other hand, the integration of logic and probability theory has been the subject of numerous studies [1, 5, 7, 8, 15, 22, 19]. While [16, 17] provides more details on these works, it suffices to point out here that these works have concerns quite different from ours, and that it is unclear how to use these formalisms to support probabilistic logic programs and deductive databases.

## 7 Conclusions

In this paper we study the semantics and the uses of probabilistic logic programs with non-monotonic negation (i.e. gp-programs). Based on the stable semantical approach for classical logic programming, we investigate the notion of stable formula functions. We show that stable formula functions are minimal fixpoints of operators associated with gp-programs. While some gp-programs may not have stable formula functions, we provide a stable class semantics that applies to all gp-programs. Finally, we demonstrate by examples how the proposed semantics can handle default reasoning appropriately in the context of probabilistic deduction.

In ongoing research, we are studying how to support empirical probabilities in our framework. We are also interested in designing a proof procedure for gp-programs. In particular, we are investigating whether it suffices to augment the proof procedure we developed for positive probabilistic logic programs with some kind of negation as failure rule.

### Acknowledgements

This research was partially sponsored by the National Science Foundation under Grant IRI-8719458 and by the "Office of Graduate Studies and Research of the University of Maryland."